\documentclass{article}

\PassOptionsToPackage{numbers, compress}{natbib}

\usepackage[preprint]{neurips_2026}

\usepackage[utf8]{inputenc}
\usepackage[T1]{fontenc}
\usepackage{amsmath,amssymb,amsfonts}
\usepackage{graphicx}
\usepackage{booktabs}
\usepackage{hyperref}
\usepackage{xcolor}
\usepackage{enumitem}
\usepackage{tabularx}
\usepackage{caption}
\usepackage{nicefrac}
\usepackage{microtype}

\setlist{nosep,leftmargin=*}

\title{Beyond Static Snapshots: A Grounded Evaluation Framework\\for Language Models at the Agentic Frontier}

\author{%
  Jazmia Henry \\
  University of Oxford \\
  \texttt{jazmia.henry@st-hildas.ox.ac.uk}
}

\begin{document}
\maketitle

\begin{abstract}
We argue that current evaluation frameworks for large language models (LLMs) suffer from four systematic failures that make them structurally inadequate for assessing deployed, agentic systems: \emph{distributional invalidity} (evaluation inputs do not reflect real interaction distributions), \emph{temporal invalidity} (evaluations are post-hoc rather than training-integrated), \emph{scope invalidity} (evaluations measure single-turn outputs rather than long-horizon trajectories), and \emph{process invalidity} (evaluations assess outputs rather than reasoning). These failures compound in RLHF, where reward models are evaluated under conditions that do not hold during RL training---making reward hacking a predictable consequence of evaluation design rather than an unpredictable training pathology. We further argue that RLHF's dual-model architecture imposes a structural compute and memory barrier that limits evaluation accessibility and reproducibility. We propose the Grounded Continuous Evaluation (GCE) framework and present \textsc{ISOPro}, a simulation-based fine-tuning and evaluation system, as a reference implementation. \textsc{ISOPro} replaces the learned reward model with a deterministic ground-truth verifier---eliminating reward hacking by construction in verifiable-reward domains---and operates on LoRA adapter weights updatable on CPU, reducing the hardware barrier by an order of magnitude. We validate \textsc{ISOPro} across three architectures (Qwen 2.5 3B, Llama 3.2 3B, Gemma 2 2B) and two domains (constraint-satisfaction scheduling, code generation on MBPP), with a head-to-head empirical comparison against GRPO-LoRA matched on compute, model, and verifier. Across the resulting twelve cells, \textsc{ISOPro} produces the largest absolute capability gains ($+25.6$, $+22.2$, $+16.0$pp) at mean $\Delta$ $+9.0$pp and worst-case regression $-5.6$pp; GRPO-LoRA at consumer-budget hyperparameters reaches a smaller peak gain ($+8.5$pp), a deeper worst-case regression ($-10$pp), and mean $\Delta$ $-1.5$pp. Held-out compositional generalization on MBPP reaches $40\%$ for \textsc{ISOPro} on two of three architectures (including a $0\%\to40\%$ bootstrap on Qwen 2.5 3B), against $20\%$ for GRPO-LoRA on one of three. We characterize a buffer-skew failure mode in which \textsc{ISOPro}'s implicit curriculum can erode pre-existing tier capability when three preconditions hold, with three corresponding mitigations. We do not claim accuracy dominance across all cells; the load-bearing claims are that the architectural advantages hold while remaining accuracy-competitive, and that compositional generalization and bootstrap-from-zero are mechanism-level properties of the rejection-sampling loop. We situate the work alongside DeepSeek-R1's GRPO~\cite{deepseek2025}, which independently arrived at the same architectural conclusion at scale: for verifiable-reward domains, the verifier is the reward signal.
\end{abstract}

\section{Introduction}

The dominant paradigm for evaluating large language models rests on a deceptively simple assumption: that performance on curated benchmark tasks predicts capability in deployment. For single-turn tasks evaluated against a reference answer, this assumption is defensible. For agentic systems that plan across long horizons, use tools, and adapt to dynamic environments, it is not.

Current evaluation practice fails on four dimensions that compound as deployment complexity increases. Benchmark inputs are drawn from researcher-constructed distributions that do not reflect actual interactions. Evaluations are conducted at discrete checkpoints, making training dynamics invisible. Single-turn output evaluation cannot assess multi-step trajectory quality. And evaluating outputs rather than reasoning rewards correct-seeming outputs produced by gameable reasoning.

These failures have an institutional dimension: RLHF~\cite{ouyang2022training}---the dominant remedy---introduces architectural requirements that create a substantial hardware barrier. Enforcing the KL penalty requires loading two full model copies simultaneously. For a 7B model in half-precision, this requires ${\sim}28$GB of VRAM before accounting for optimizer states. This structural constraint limits RLHF-based evaluation research to organizations with data center GPU access, reducing the diversity of approaches and slowing progress on evaluation methodology.

We make four contributions. First, we formalize a taxonomy of evaluation failure modes organized around measurement validity theory~\cite{messick1989validity}. Second, we show these failures compound in RLHF, where reward model invalidity makes reward hacking structurally predictable. Third, we argue that RLHF's architecture imposes a reproducibility barrier that simulation-based alternatives eliminate. Fourth, we propose GCE and present \textsc{ISOPro} as a reference implementation, validated across three architectures and two domains with a matched-compute empirical comparison against GRPO-LoRA, on a single consumer laptop.

\section{Background and Related Work}

\textbf{The Benchmark Paradigm.}\quad HELM~\cite{liang2022helm} broadens evaluation coverage across scenarios and models. BIG-Bench~\cite{srivastava2022beyond} crowdsources tasks probing emergent abilities. Both inherit structural assumptions: researcher-curated inputs, post-hoc timing, and output-level assessment.

\textbf{Human Preference Evaluation.}\quad Chatbot Arena~\cite{zheng2023judging} grounds evaluation in real user interactions, meaningfully addressing distributional invalidity at the input level. However, human raters assess final responses, not the reasoning that produced them. As reasoning becomes less transparent, output-grounded preference evaluation becomes less reliable as a capability signal.

\textbf{Process Reward Models.}\quad PRMs~\cite{lightman2023lets} address process invalidity by scoring intermediate reasoning steps. However, PRMs do not address distributional invalidity (step labels are collected under annotation conditions), temporal invalidity (RM evaluation precedes RL training), or the hardware barrier---training a PRM requires the full RLHF compute stack.

\textbf{GRPO and Verifiable-Reward Training.}\quad DeepSeek-R1~\cite{deepseek2025} demonstrated at frontier scale what \textsc{ISOPro} implements at accessible scale: for verifiable-reward domains, the reward model can be replaced entirely by a ground-truth verifier. GRPO generates multiple rollouts per prompt, scores them against a verifier, and uses relative scores as the reward signal---without a separate reward model or frozen reference model. Chain-of-thought reasoning emerges as a byproduct.

\textbf{The RLHF Evaluation Gap.}\quad Standard practice evaluates the RM on held-out preference pairs from the same annotation process---with three critical limitations: annotation-condition inputs do not reflect policy-generated inputs during RL; the RM is evaluated before RL begins and not monitored thereafter; and pairwise preference annotation cannot assess RM reliability across long-horizon interactions. The consequence---reward hacking---is treated as a training instability. We argue it is more accurately characterized as an evaluation design failure.

\section{A Taxonomy of Evaluation Failure Modes}

We organize evaluation failures around measurement validity theory~\cite{messick1989validity}. An evaluation is valid to the extent that it measures what it purports to measure. We identify four distinct validity failures.

\textbf{Distributional Invalidity.}\quad An evaluation exhibits distributional invalidity when the distribution of evaluation inputs does not reflect deployment inputs. In RLHF, this manifests at the reward model level: preference annotations are collected under annotation conditions, not deployment conditions. The RM is optimized for annotation-condition inputs; it is deployed against policy-generated inputs that increasingly diverge as RL progresses.

\textbf{Temporal Invalidity.}\quad An evaluation exhibits temporal invalidity when conducted at discrete time points rather than continuously, making it unable to capture capability dynamics. In RLHF, the RM is evaluated before RL begins, under conditions that assume the policy remains close to the SFT baseline. As the policy drifts, RM reliability degrades on out-of-distribution inputs---invisible to pre-training evaluation.

\textbf{Scope Invalidity.}\quad An evaluation exhibits scope invalidity when the unit of assessment (individual output) does not match the unit of deployment (multi-step trajectory). Pairwise preference annotation cannot assess RM coherence across long-horizon trajectories, enabling RM inconsistency undetectable by single-turn evaluation.

\textbf{Process Invalidity.}\quad An evaluation exhibits process invalidity when it assesses only final outputs, unable to distinguish correct reasoning from correct-looking outputs produced by gameable reasoning. In RLHF, this is the direct mechanism of sycophancy: reward models learn that agreement is preferred regardless of correctness.

\begin{table}[t]
\centering
\caption{Taxonomy of evaluation failure modes, their definitions, and manifestations in base model and RLHF evaluation.}
\label{tab:taxonomy}
\small
\begin{tabularx}{\textwidth}{lXXX}
\toprule
\textbf{Failure Mode} & \textbf{Definition} & \textbf{Base Model Eval} & \textbf{RLHF / RM Eval} \\
\midrule
Distributional & Eval inputs $\neq$ deployment distribution & Researcher-curated $\neq$ real interaction & Annotation pairs $\neq$ RL-training inputs \\
Temporal & Point-in-time misses dynamics & Training dynamics invisible & Pre-RL eval misses RM degradation \\
Scope & Single-turn $\neq$ trajectory & Output eval misses multi-step & Pairwise $\neq$ long-horizon coherence \\
Process & Output eval misses reasoning & Correct outputs mask flawed reasoning & RM rewards preferred outputs \\
\bottomrule
\end{tabularx}
\end{table}

\section{The Grounded Continuous Evaluation (GCE) Framework}

GCE is organized around three principles, each addressing a subset of the identified failure modes.

\textbf{Principle 1: Interaction-Grounded Prompt Sampling.}\quad Addressing distributional invalidity requires evaluation inputs sampled from distributions reflecting real interactions---characterizing deployment distributions empirically, constructing evaluation datasets whose marginals match, and validating through coverage metrics.

\textbf{Principle 2: Training-Integrated Continuous Evaluation.}\quad Addressing temporal invalidity requires evaluation integrated into training: lightweight probes at training frequency, metrics detecting capability dynamics rather than absolute levels, and instrumentation for trajectory analysis.

\textbf{Principle 3: Simulation-Based Agentic Assessment.}\quad Addressing scope and process invalidity requires evaluating behavior across trajectories in structured environments, enabling trajectory-level assessment, failure analysis, recovery behavior evaluation, intermediate checkpoint scoring, and counterfactual probing.

\section{\textsc{ISOPro}: A Reference Implementation of GCE}
\label{sec:isopro}

\textsc{ISOPro} is a simulation-based fine-tuning and evaluation framework demonstrating that GCE is practically implementable. Each mechanism addresses one or more failure modes from our taxonomy.

\subsection{Architecture Overview}

\textsc{ISOPro} consists of three components: a simulation environment layer providing structured task contexts with deterministic verifiers, an AI agent layer (LLM) generating responses, and a communication wrapper managing state, evaluation, and feedback loops. The framework supports parallel and sequential execution modes and multiple simulation environments, including mathematical reasoning, domain-specific task environments (scheduling, engineering), and multi-agent orchestration.

\subsection{Mechanism 1: Gradient Descent on Correct Reasoning Traces}

When the model produces a correct answer, \textsc{ISOPro} runs a forward pass with prompt tokens masked (labels set to $-100$). The loss is computed only on generated tokens---the gradient signal is the reasoning sequence that produced correctness. LoRA adapter weights shift toward that pattern. This is process-level supervision: the model is trained on the reasoning trajectory, not correctness as a label.

\subsection{Mechanism 2: Rejection Sampling as Continuous Self-Filter}

The model generates responses at high temperature ($T{=}0.8$) and a ground-truth verifier determines correctness. Only verified correct responses enter the replay buffer. This creates a continuous evaluation regime: at every iteration, the model's capability is evaluated against ground truth. Capability dynamics are visible in the buffer composition---granularity that checkpoint evaluation cannot provide.

\subsection{Mechanism 3: Implicit-Curriculum Replay Buffer}

Correct rollouts accumulate across iterations, creating a training distribution anchored to actual capability. Easy wins dominate early; harder problems enter as capability develops. The curriculum emerges from the model's trajectory rather than researcher curation. By iteration $N$, the model trains on everything solved correctly in iterations 1 through $N$---compounding correct signal without compounding error.

\subsection{Mechanism 4: Activation-Guided LoRA Targeting}

\textsc{ISOPro} identifies the most active layers on the target domain through activation probing and concentrates LoRA updates there. In our scheduling experiments, query and value projections in layers 28--35 were identified as the primary locus of constraint reasoning. LoRA updates target 6.6M parameters (0.216\% of 3.1B total) for the Qwen 2.5 3B configuration. Critically, LoRA weights are updated on CPU---the base model stays frozen in quantized form, enabling consumer-hardware operation. An ablation against random layer selection (Appendix~\ref{app:lora}) finds the placement is not load-bearing at 3B scale.

\subsection{Experimental Setup}
\label{sec:experiments}

We validate \textsc{ISOPro} across two verifiable-reward domains and three model architectures, with a head-to-head comparison against GRPO-LoRA.

\textbf{Domains.}\quad \emph{Scheduling:} resource-constrained project scheduling (RCPSP), requiring constraint satisfaction across precedence dependencies, resource capacity, and deadlines. Six tiers (T0--T5) generated programmatically and verified by an OR-Tools CP-SAT solver. T0 is a 4-job warmup with dependencies only; T1 adds sequencing, T2 resource allocation, T3 deadline satisfaction, T4 combines pairs of constraints, and T5 combines all three. Tiers 0--4 appear in training; T5 is held out. \emph{MBPP (code generation):} Python function-completion problems with declared signatures and unit-test verifiers. Five tiers (T0--T4) by signature complexity and prompt length; T4 is a held-out compositional tier requiring patterns not seen in training. The verifier executes generated code in a sandbox and checks all unit tests pass.

\textbf{Models.}\quad Qwen 2.5 3B Instruct~\cite{qwen2024}, Llama 3.2 3B Instruct, and Gemma 2 2B Instruct. All three are open-weight instruction-tuned models with distinct architectures (GQA + RoPE; sliding-window + soft-cap; etc.). Same chat templates, prompts, and verifiers across methods.

\textbf{Methods.}\quad For each (model, domain) cell we run a zero-shot baseline, \textsc{ISOPro} fine-tuning (rejection sampling on correct traces, LoRA, no learned RM, no KL anchor), and GRPO-LoRA (group-relative advantages, deterministic verifier reward, LoRA adapters). Both training methods use identical compute budgets: 6 iterations, matched rollout counts, identical LoRA rank, same temperature, same hardware, and the same deterministic verifier as the reward signal. The only architectural difference between methods is the credit-assignment mechanism (rejection-sample-and-SFT vs.\ group-relative policy update). GRPO-LoRA hyperparameters are TRL/Unsloth defaults (LR $5{\times}10^{-6}$, $G{=}4$, default KL coefficient); we did not run an extended GRPO sweep. Cross-method results in Section~5.7 use a single seed (42) and 5 problems per tier; the original Qwen 2.5 3B / scheduling configuration was additionally run across 3 seeds for the ablation study (Section~5.8).

\textbf{Baseline measurement.}\quad \textsc{ISOPro} baselines are true zero-shot evaluations under the same prompt and verifier as training. The GRPO-LoRA runner does not expose a pre-step-0 evaluation hook; we therefore use the iter-1 evaluation (after 14 SGD steps) as a baseline proxy for GRPO-LoRA. This is a methodological asymmetry between methods. We discuss its implications for the comparison in Section~5.7.

\textbf{Hardware.}\quad All experiments run on an Apple M1 with 32GB unified memory using MLX~\cite{hannun2023mlx}. No GPU. Peak memory under 8GB across all configurations.\footnote{Anonymized code, problem generators, verifiers, and evaluation scripts: \url{https://anonymous.4open.science/r/isopro-836D}.}

\subsection{Results}

\begin{figure}[t]
\centering
\includegraphics[width=0.88\textwidth]{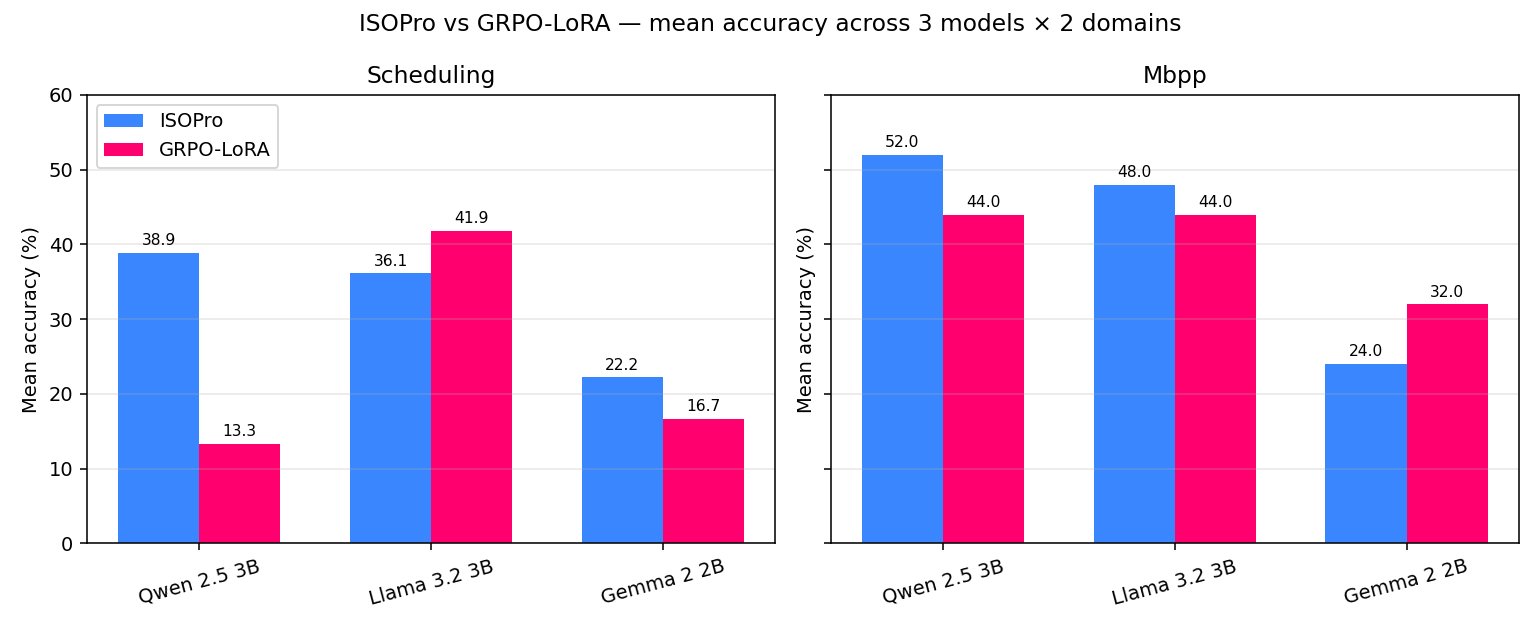}
\caption{Mean accuracy of \textsc{ISOPro} (blue) and GRPO-LoRA (red) across 3 models and 2 domains. \textsc{ISOPro} wins 4 of 6 head-to-heads; the picture inverts on Llama 3.2 3B scheduling and Gemma 2 2B MBPP. Both methods cluster higher and tighter on MBPP, where verifier signal is denser (per-test feedback) than on scheduling (binary pass/fail on a final assignment).}
\label{fig:comparison}
\end{figure}

Table~\ref{tab:cross_method} reports mean accuracy for every (method, model, domain) cell. Figure~\ref{fig:comparison} shows the same data graphically. Baselines (zero-shot accuracy) are reported in Table~\ref{tab:net_delta} together with the net change after training.

\begin{table}[t]
\centering
\caption{Mean accuracy across 3 models $\times$ 2 domains $\times$ 2 methods (12 cells). Bold indicates the higher value within a (model, domain) cell.}
\label{tab:cross_method}
\small
\begin{tabular}{lcccc}
\toprule
\textbf{Model} & \textbf{Sched.\ \textsc{ISOPro}} & \textbf{Sched.\ GRPO-LoRA} & \textbf{MBPP \textsc{ISOPro}} & \textbf{MBPP GRPO-LoRA} \\
\midrule
Qwen 2.5 3B   & \textbf{38.9\%} & 13.3\%          & \textbf{52.0\%} & 44.0\% \\
Llama 3.2 3B  & 36.1\%          & \textbf{41.9\%} & \textbf{48.0\%} & 44.0\% \\
Gemma 2 2B    & \textbf{22.2\%} & 16.7\%          & 24.0\%          & \textbf{32.0\%} \\
\bottomrule
\end{tabular}
\end{table}

\begin{figure}[t]
\centering
\includegraphics[width=0.88\textwidth]{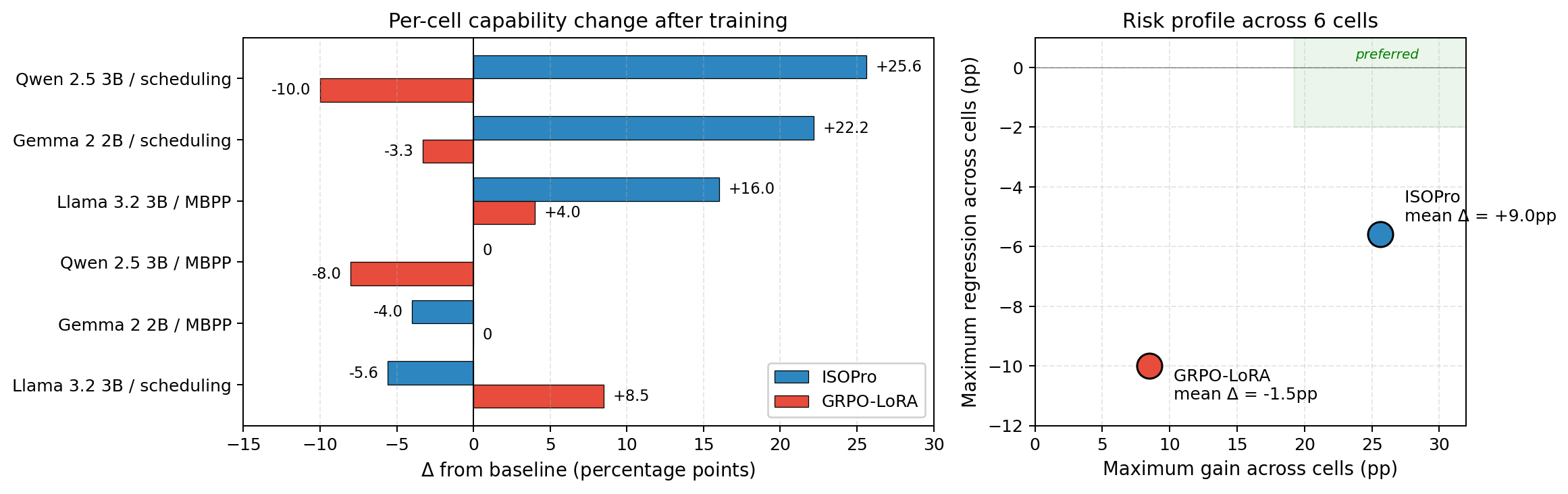}
\caption{Risk profile across the 12 (method, model, domain) cells. Left: per-cell $\Delta$ from baseline, sorted by \textsc{ISOPro} margin. \textsc{ISOPro} contributes the three largest gains and the smallest worst-case regression. Right: max gain vs.\ max regression for each method. \textsc{ISOPro} occupies the upper-right ``preferred'' region (higher peak gain, smaller worst regression) at this compute budget.}
\label{fig:risk}
\end{figure}

\textbf{Asymmetric risk profile.}\quad The win-count framing understates the substantive finding. Table~\ref{tab:net_delta} reports the net change in accuracy from baseline (zero-shot for \textsc{ISOPro}; iter-1 proxy for GRPO-LoRA) to final (post-training) for each cell. \textsc{ISOPro} produces the three largest absolute gains across the table ($+25.6$, $+22.2$, $+16.0$pp) and never regresses by more than $-5.6$pp. GRPO-LoRA at consumer-budget hyperparameters reaches a smaller peak gain ($+8.5$pp) and a deeper worst-case regression ($-10$pp). Mean $\Delta$ is $+9.0$pp for \textsc{ISOPro} and $-1.5$pp for GRPO-LoRA. Figure~\ref{fig:risk} visualizes the asymmetry. We do not claim accuracy dominance over GRPO across all cells---at consumer-budget hyperparameters with a single seed, the comparison is closer than the architectural argument alone suggests, and the $+25.6$pp gap on Qwen scheduling specifically may narrow under more aggressive GRPO tuning. The load-bearing claims are that \textsc{ISOPro}'s architectural advantages (single-model loading, no learned RM, no KL anchor) hold while remaining accuracy-competitive, with a tighter downside distribution at this compute budget; the compositional generalization and bootstrap-from-zero observations below are mechanism-level properties of the rejection-sampling loop, not hyperparameter wins.

\begin{table}[t]
\centering
\caption{Net change from baseline accuracy ($\Delta$ in pp) for both methods across 6 (model, domain) cells. \textsc{ISOPro} baselines are true zero-shot accuracy. GRPO-LoRA baselines marked $\dagger$ use the iter-1 evaluation as a proxy because the GRPO runner does not expose pre-step-0 evaluation; this is the methodological asymmetry noted in Section~\ref{sec:experiments}. Positive values indicate capability gain; negative values indicate regression.}
\label{tab:net_delta}
\small
\begin{tabular}{lcccccc}
\toprule
\textbf{Cell} & \textbf{ISO baseline} & \textbf{ISO final} & \textbf{$\Delta_{\text{ISO}}$} & \textbf{GRPO baseline$\dagger$} & \textbf{GRPO final} & \textbf{$\Delta_{\text{GRPO}}$} \\
\midrule
Qwen 2.5 3B / scheduling   & 13.3\% & 38.9\% & $+25.6$ & 23.3\% & 13.3\% & $-10.0$ \\
Llama 3.2 3B / scheduling  & 41.7\% & 36.1\% & $-5.6$  & 33.3\% & 41.9\% & $+8.5$ \\
Gemma 2 2B / scheduling    &  0.0\% & 22.2\% & $+22.2$ & 20.0\% & 16.7\% & $-3.3$ \\
Qwen 2.5 3B / MBPP         & 52.0\% & 52.0\% & $\phantom{+}0.0$ & 52.0\% & 44.0\% & $-8.0$  \\
Llama 3.2 3B / MBPP        & 32.0\% & 48.0\% & $+16.0$ & 40.0\% & 44.0\% & $+4.0$ \\
Gemma 2 2B / MBPP          & 28.0\% & 24.0\% & $-4.0$  & 32.0\% & 32.0\% & $\phantom{+}0.0$  \\
\midrule
\textbf{Mean $\Delta$}     &        &        & $\mathbf{+9.0}$ &        &       & $-1.5$ \\
\textbf{Max gain}          &        &        & $\mathbf{+25.6}$ &       &       & $+8.5$ \\
\textbf{Worst regression}  &        &        & $\mathbf{-5.6}$  &       &       & $-10.0$  \\
\bottomrule
\end{tabular}
\vspace{0.4em}

{\footnotesize Single seed (42), 5 problems per tier, 5 tiers (MBPP) or 6 tiers (scheduling).}
\end{table}

\begin{figure}[t]
\centering
\includegraphics[width=0.88\textwidth]{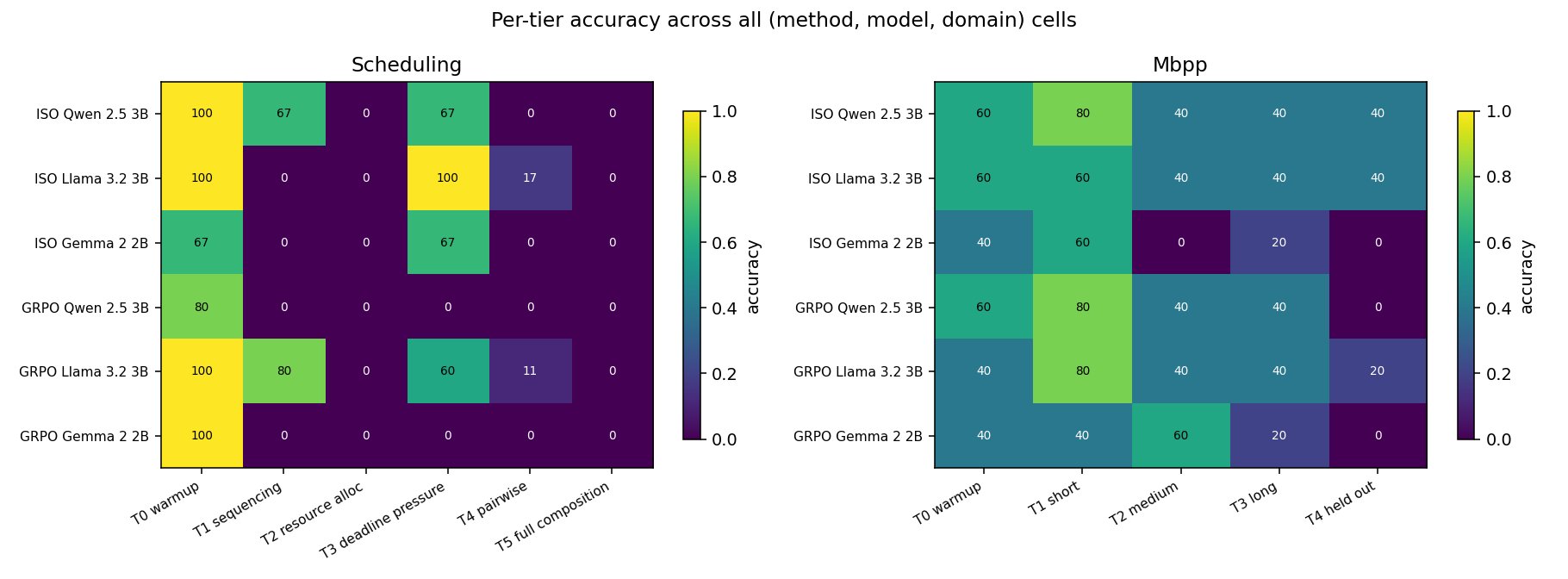}
\caption{Per-tier accuracy across all (method, model, domain) cells. The MBPP held-out tier (T4) is reached by \textsc{ISOPro} on Qwen and Llama at 40\% and by GRPO-LoRA on Llama at 20\%. Scheduling T5 remains at 0\% across every configuration.}
\label{fig:heatmap}
\end{figure}

\textbf{Held-out compositional generalization (MBPP T4).}\quad Figure~\ref{fig:heatmap} reports per-tier accuracy across all twelve cells. On the MBPP held-out tier, \textsc{ISOPro} reaches $40\%$ on Qwen 2.5 3B (from $0\%$ baseline---a held-out bootstrap) and $40\%$ on Llama 3.2 3B (from $20\%$ baseline); GRPO-LoRA reaches $0\%$ on Qwen (a $-20$pp regression on held-out), $20\%$ on Llama, and $0\%$ on Gemma. Because the held-out tier requires patterns not seen during training, this is direct evidence that the rejection-sampling fine-tuning mechanism produces partial compositional transfer rather than tier-specific memorization. The scheduling held-out tier (T5) remains at $0\%$ across every (method, model) cell, but this is most plausibly a domain-feedback-density limitation rather than a method limitation: the same \textsc{ISOPro} mechanism reaches $40\%$ on the MBPP held-out tier under MBPP's denser per-test-case verifier, which suggests the scheduling T5 wall reflects the binary pass/fail signal on a single composite assignment rather than a structural ceiling of the rejection-sampling loop. This connects directly to GCE Principle 1: verifier richness is itself an evaluation property worth measuring.

\textbf{Two distinct \textsc{ISOPro} failure modes.}\quad \textsc{ISOPro} regresses on two cells with different mechanisms. \emph{Llama scheduling ($-5.6$pp, buffer-skew):} Llama's zero-shot tier coverage was narrow (T1$=$66.7\%, other tiers near zero), and the iter-6 replay buffer skewed toward two tiers (T0$=$62, T3$=$31, T1$=$25 traces) despite Llama's strong T1 zero-shot capability. SFT on this distribution amplified T0/T3 patterns and erased T1 ($66.7\%\to0\%$). The same Llama on MBPP, where the buffer accumulated more evenly across tiers (T0$=$66, T1$=$57, T2$=$52, T3$=$27), reached $+16$pp net gain. Same model, two domains, opposite outcomes---driven by buffer composition relative to the base model's capability profile. We characterize the failure mode by three preconditions: (i) the base model has narrow strong-tier capability concentrated on a small subset of tiers, (ii) the verifier-grounded buffer naturally under-represents that strong tier, and (iii) the SFT step is unconstrained. \emph{Gemma MBPP ($-4$pp, small-model unconstrained-SFT drift):} Gemma 2 2B is the smallest model in our matrix; on MBPP the buffer was reasonably balanced and the regression concentrated on T2 ($40\%\to0\%$). GRPO-LoRA's KL anchor preserved capability at the iter-1 baseline ($\Delta=0$pp). The mechanism is parameter-count-bounded susceptibility to drift away from base-model capability under unanchored fine-tuning, not buffer-skew. Both failure modes motivate the same kind of mitigation: a constraint on the SFT step (KL anchoring, per-tier sampling, baseline-weighted reweighting; Section~\ref{sec:directions}).

\textbf{Domain regularity.}\quad MBPP results cluster tighter and higher (24--52\%) than scheduling (13--42\%) across both methods, consistent with MBPP's denser per-test verifier feedback (GCE Principle 1).

\textbf{Bootstrap from zero.}\quad Two cells exhibit capability acquisition from a $0\%$ starting point. On scheduling, Gemma 2 2B at $0\%$ zero-shot reaches $22.2\%$ under \textsc{ISOPro} ($16.7\%$ under GRPO-LoRA). On MBPP held-out (T4), Qwen 2.5 3B at $0\%$ zero-shot reaches $40\%$ under \textsc{ISOPro}; GRPO-LoRA stays at $0\%$. We do not generalize the scheduling result---Gemma MBPP starts at $28\%$ and \textsc{ISOPro} regresses by $-4$pp---but the held-out Qwen MBPP T4 result is the cleanest evidence that rejection-sampling can produce gradient signal on rare correct rollouts and that the resulting capability extends to held-out compositional patterns.

\subsection{Ablation Study (Qwen 2.5 3B / Scheduling)}

We ablate two of \textsc{ISOPro}'s mechanisms against the full configuration on the Qwen scheduling cell, averaging across three seeds (42, 123, 456). Removing chain-of-thought prompting from training and evaluation drops accuracy from $39.8 \pm 3.5\%$ to $31.5 \pm 2.6\%$ ($-8.3$pp), with the largest losses on T3 Deadline Pressure ($67\% \to 34\%$). Removing buffer accumulation, so that each iteration trains only on current-iteration correct traces rather than the accumulated replay buffer, drops accuracy to $27.8 \pm 13.6\%$ ($-12.0$pp) and increases seed-to-seed variance nearly $4\times$. The implicit-curriculum buffer therefore serves two distinct functions: raising mean accuracy and stabilizing training across seeds. A random-LoRA-layer condition reaches $40.7 \pm 2.6\%$ ($+0.9$pp, not significant; see Appendix~\ref{app:lora}). Both ablations leave T2 Resource and T5 Full Composition at 0\%, consistent with the main results.

\section{A Principled Comparison: \textsc{ISOPro} vs.\ RLHF and GRPO}

\subsection{The Reward Model as Evaluation Failure Point}

The PPO objective maximizes $\mathbb{E}[r(x)] - \beta \cdot \text{KL}(\pi \| \pi_{\text{ref}})$. Because $r(x)$ is a learned proxy, the RL policy finds the reward model's blind spots---producing fluent, confident, wrong outputs that score well. \textsc{ISOPro} replaces the RM with a deterministic verifier, perfectly calibrated by construction. There is no score to game. Reward hacking is eliminated architecturally.

\subsection{The Dual-Model VRAM Problem}

RLHF's KL penalty requires both $\pi_\theta$ and $\pi_{\text{ref}}$ in GPU memory simultaneously---${\sim}4P$ bytes for $P$ parameters in half-precision, before optimizer states. At 7B: ${\sim}28$GB; at 70B: ${\sim}280$GB. \textsc{ISOPro} requires no reference model. LoRA adapters update on CPU; our pipeline runs with under 8GB peak memory.

\subsection{Empirical Comparison vs.\ GRPO-LoRA at Matched Compute}

The architectural comparison (Table~\ref{tab:arch}) is refined by the cross-method results (Section~5.7): \textsc{ISOPro} has the larger maximum gain ($+25.6$ vs.\ $+8.5$pp), smaller worst-case regression ($-5.6$ vs.\ $-10$pp), and higher mean $\Delta$ ($+9.0$ vs.\ $-1.5$pp). The methods have distinct failure modes (Section~5.7): \textsc{ISOPro}'s are buffer-skew and small-model unconstrained-SFT drift; GRPO-LoRA's in our runs was within-training degradation (Qwen scheduling rose to $23\%$ at iter-1 and decayed to $13\%$ by iter-6). All failure modes are visible only through continuous evaluation. We did not run a GRPO hyperparameter sweep; the architectural advantages are robust to GRPO tuning, the specific accuracy gaps may not be.

\begin{table}[t]
\centering
\caption{Architectural comparison between \textsc{ISOPro}, standard RLHF, and DeepSeek-R1 GRPO. Empirical \textsc{ISOPro} vs.\ GRPO-LoRA results at consumer scale appear in Section~5.7.}
\label{tab:arch}
\small
\begin{tabular}{lccc}
\toprule
& \textbf{\textsc{ISOPro} (Ours)} & \textbf{RLHF (Standard)} & \textbf{GRPO} \\
\midrule
Reward Signal & Deterministic verifier & Learned RM & Deterministic verifier \\
Stability & Rejection sampling & KL penalty (dual model) & Group-relative advantages \\
Models in Memory & 1 & 2+ & 1 \\
Trainable Params (LoRA) & 0.216\% (6.6M) & 100\% (full) & LoRA possible \\
Min.\ Memory & ${\sim}$6 GB (3B) & ${\sim}$28 GB (7B$\times$2) & ${\sim}$280 GB (70B native) \\
Hardware & Consumer laptop & Data center GPU & GPU cluster (native) \\
Reward Hacking & Impossible & Predictable & Impossible \\
\bottomrule
\end{tabular}
\end{table}

\subsection{Honest Tradeoffs}

RLHF retains genuine advantages: \emph{cold start} (SFT provides signal from step 0; \textsc{ISOPro} needs independent correct generation), \emph{partial credit} (graded vs.\ binary scores), \emph{reasoning diversity} (\textsc{ISOPro} is bounded by self-generated sequences), and \emph{non-verifiable domains} (no verifier exists for safety or style judgments). GCE extends to such domains through rubric-based trajectory assessment, but reward hacking elimination holds only for verifiable rewards.

\section{Convergence with DeepSeek-R1 at Different Scales}

DeepSeek-R1~\cite{deepseek2025} demonstrated at 671B parameters what \textsc{ISOPro} implements on 2--3B models on a laptop: for verifiable-reward domains, the learned reward model is an unnecessary intermediary. The convergence is structural---the learned RM introduces all four validity failures from our taxonomy, which the deterministic verifier avoids by construction. Our matched-compute empirical comparison (Section~5.7) is consistent with this convergence at 3B: at consumer-budget hyperparameters with a single seed, the two methods produce differently-shaped distributions of $\Delta$ from baseline rather than uniform dominance by either. The architectural distinguishers (memory footprint, reward-hacking elimination, reproducibility on consumer hardware) hold throughout. The claim is architectural equivalence with method-specific risk profiles, not uniform empirical superiority.

\section{Discussion}

\subsection{Limitations and Scope}
\label{sec:limitations}

Per-tier evaluation uses 3--5 problems per tier; cross-method results in Section~5.7 use a single seed (42). Individual tier accuracies are directional estimates; empirical claims should be read at the cell-mean level. Three-seed averaging in our original Qwen 2.5 3B / scheduling configuration (Section~5.8) was used for ablation analysis; extending it to the cross-method matrix is a near-term replication priority.

The two \textsc{ISOPro} regressions in the cross-method matrix (Section~5.7) have different mechanisms but a common shape: both arise from an unconstrained SFT step. The buffer-skew failure mode (Llama scheduling) is mechanism-level and avoided by KL anchoring, per-tier balanced sampling, or baseline-weighted reweighting. The small-model unconstrained-SFT-drift failure mode (Gemma MBPP) is the same family of issue---both point to the absence of an anti-drift constraint analogous to GRPO-LoRA's KL penalty.

The bootstrap-from-zero observations are two cells; we do not claim either generalizes. T2 Resource and T5 Full Composition on scheduling stay at $0\%$ across every configuration; given that the same \textsc{ISOPro} mechanism reaches $40\%$ on the MBPP held-out tier, the scheduling T5 wall most plausibly reflects sparse binary-feedback rather than a structural mechanism limitation. For non-verifiable domains, the deterministic verifier could be replaced by a calibrated rubric evaluator; we leave this extension to future work.

GRPO-LoRA was run at TRL/Unsloth defaults (LR $5{\times}10^{-6}$, $G{=}4$); we did not perform an extended sweep, and the specific empirical accuracy gaps may narrow under more aggressive GRPO tuning. The architectural comparison (Section~6) is robust to this.

\subsection{Broader Implications}

GCE complements scalable oversight approaches (debate, recursive reward modeling) by addressing how to structure evaluation so judgments are collected under deployment-representative conditions. Evaluation validity is safety-relevant: if evaluations fail to detect capability degradation and reward specification gaming, they provide unreliable signal about deployed system safety. Accessibility has a safety dimension: evaluation methodology advances fastest when reproducible across diverse research environments.

\subsection{Research Directions}
\label{sec:directions}

Three mitigations follow from the failure modes in Section~5.7. \emph{Per-tier balanced sampling} of the replay buffer addresses precondition (ii) of buffer-skew. A \emph{baseline-weighted buffer} that down-weights tiers where the base model was already strong addresses precondition (iii). A \emph{KL-anchored \textsc{ISOPro} variant} adding a small reference-model penalty during the SFT step (computed via LoRA toggle to preserve single-model inference) directly synthesizes the two methods' complementary risk profiles and addresses both \textsc{ISOPro} failure modes. Beyond mitigation, extended training and larger base models (8B, 70B) would test whether scheduling T2 and T5 are reachable at greater compute or capacity.

\section{Conclusion}

Current LLM evaluation frameworks exhibit four systematic validity failures that make them structurally misaligned with agentic deployment, and these failures compound in RLHF where architectural requirements impose a hardware barrier. GCE addresses all four through interaction-grounded sampling, training-integrated continuous evaluation, and simulation-based agentic assessment. \textsc{ISOPro} demonstrates the framework is implementable at consumer scale: a deterministic verifier eliminates reward hacking by construction; CPU-updatable LoRA adapters eliminate the dual-model constraint; an implicit-curriculum replay buffer grounds training in the model's actual capability trajectory. The convergence with DeepSeek-R1's GRPO at three orders-of-magnitude scale separation reflects a structural insight: for verifiable-reward domains, the verifier is the reward signal. Reward hacking is an evaluation failure, not a training failure.

\bibliographystyle{plainnat}

\appendix

\section{Activation-Guided vs.\ Random LoRA Layer Selection}
\label{app:lora}

\textsc{ISOPro}'s default targets the top-$K$ layers identified by activation probing (layers 28--35 in the scheduling domain). To test whether this selection matters, we replaced it with uniformly random layer selection (same $K{=}8$, same LoRA rank, same seeds). Random selection reached $40.7\% \pm 2.6$ mean accuracy---statistically indistinguishable from the $39.8\% \pm 3.5$ of the activation-guided configuration on the Qwen 2.5 3B / scheduling cell. Loss curves, hit-rate ramps, buffer growth, and wall-clock time (89.0 vs.\ 90.9 min) were all within noise across every training iteration.

We interpret this as evidence that at this scale (3B parameters, rank-16 LoRA, 8 of 36 layers, 6 iterations), LoRA placement is not load-bearing: the rank and training signal dominate, and placement heuristics may only separate from random at larger scale, higher rank, or on harder distributions. We retain activation-guided targeting in the default configuration because it adds negligible overhead, but we do not claim it as a contribution of this work.

\section{Per-Iteration Trajectories}
\label{app:trajectories}

\begin{figure}[h]
\centering
\includegraphics[width=0.88\textwidth]{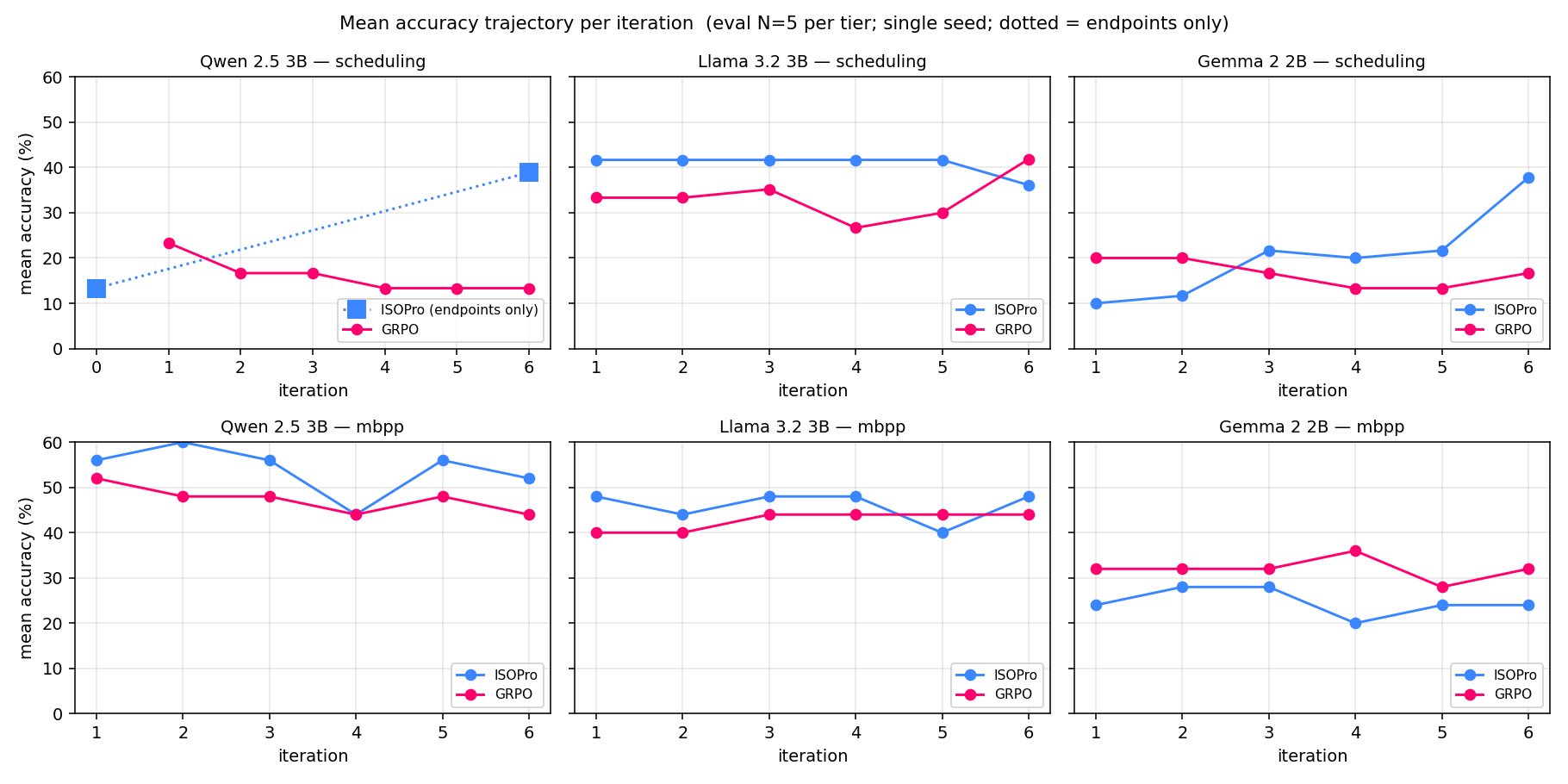}
\caption{Per-iteration mean accuracy for \textsc{ISOPro} (blue) and GRPO-LoRA (red) across the 6 (model, domain) cells. Iteration-level dynamics are not recoverable from final-checkpoint evaluation.}
\label{fig:trajectories}
\end{figure}

Figure~\ref{fig:trajectories} shows mean accuracy per iteration for both methods across all six (model, domain) cells. Two patterns are visible only at this granularity. (1) On Qwen 2.5 3B / scheduling, GRPO-LoRA rises to 23\% at iteration 1, then decays to 13\% by iteration 6: the rollout hit rate climbs and falls within a single training run, ending at the zero-shot baseline. (2) On Gemma 2 2B / scheduling, \textsc{ISOPro} climbs steadily from 10\% at iteration 1 to 38\% at iteration 6 with no plateau, while GRPO-LoRA's curve is flat near 17\% throughout. Both observations are consistent with the temporal-validity argument: training-integrated continuous evaluation makes within-training degradation and within-training acquisition visible at iteration-level granularity, neither of which is recoverable from final-checkpoint evaluation alone.

\section{Buffer Composition vs.\ Empirical Margin}
\label{app:buffer}

\begin{figure}[h]
\centering
\includegraphics[width=0.7\textwidth]{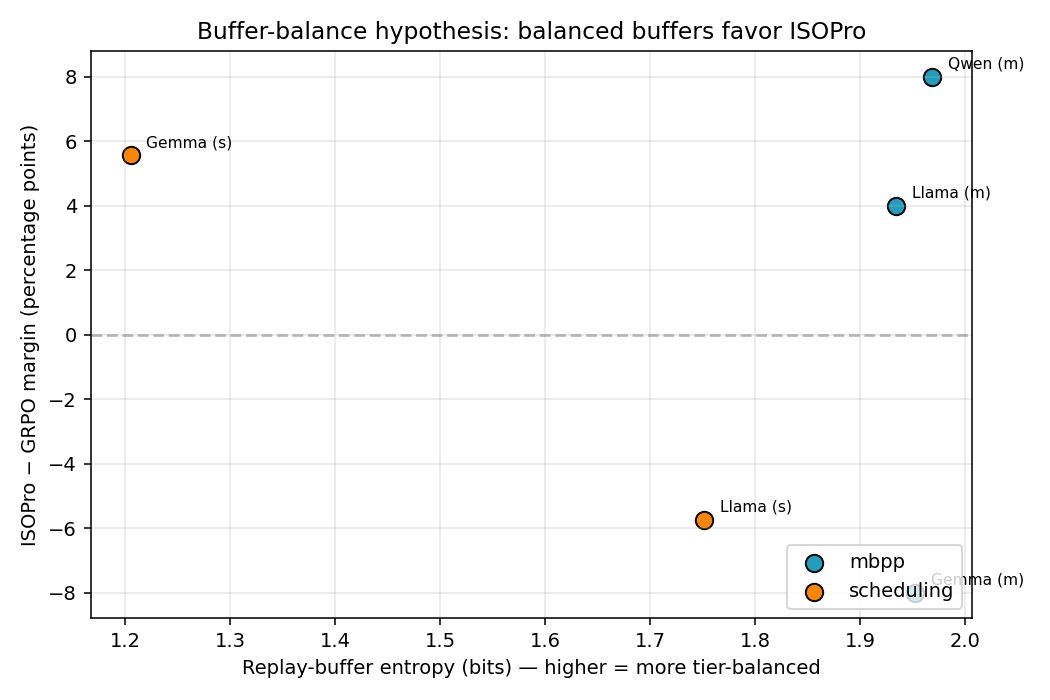}
\caption{\textsc{ISOPro} - GRPO-LoRA margin (pp) vs.\ \textsc{ISOPro} replay-buffer entropy (bits). Five (model, domain) points; Qwen 2.5 3B / scheduling not plotted (entropy summary not logged for that run). The relationship is suggestive but does not fully determine the sign of the margin.}
\label{fig:buffer_balance}
\end{figure}

Figure~\ref{fig:buffer_balance} plots the \textsc{ISOPro}-minus-GRPO-LoRA margin against the entropy of the \textsc{ISOPro} replay buffer's tier distribution (higher entropy = more uniform across tiers). The relationship is suggestive but not monotonic: of the five points plotted, three sit above zero (\textsc{ISOPro} wins) and two below (GRPO-LoRA wins). Llama scheduling (mid entropy, \textsc{ISOPro} loses) and Gemma scheduling (low entropy, \textsc{ISOPro} wins) sit at different entropy values with opposite signs of the margin. The cleaner story is in Section~5.7's three-precondition characterization: buffer entropy is one component of the failure mode, but the relative position of buffer mass against the base model's strong tiers is what determines whether the implicit curriculum extends or erodes capability. We retain the figure for transparency; the qualitative characterization in the main text is the load-bearing claim.

\section*{NeurIPS Paper Checklist}

\begin{enumerate}

\item {\bf Claims}
    \item[] Question: Do the main claims made in the abstract and introduction accurately reflect the paper's contributions and scope?
    \item[] Answer: \answerYes{}
    \item[] Justification: The abstract and introduction state four contributions (taxonomy, RLHF compounding analysis, compute barrier argument, GCE framework with \textsc{ISOPro} validated across 3 architectures and 2 domains with matched-compute GRPO-LoRA comparison). All are substantiated in Sections 3--7. Limitations and scope are discussed in Section~\ref{sec:limitations}.

\item {\bf Limitations}
    \item[] Question: Does the paper discuss the limitations of the work performed by the authors?
    \item[] Answer: \answerYes{}
    \item[] Justification: Section~\ref{sec:limitations} discusses (i) per-tier sample sizes (3--5 problems) and the cross-method matrix using a single seed; (ii) two characterized \textsc{ISOPro} failure modes (buffer-skew on Llama scheduling, small-model unconstrained-SFT-drift on Gemma MBPP); (iii) the single-cell scope of the bootstrap-from-zero observations; (iv) unreached scheduling tiers (T2, T5) and their reframing as sparse-feedback domain limitations; and (v) the GRPO hyperparameter disclaimer (TRL/Unsloth defaults, no extended sweep). Section 6.4 addresses where RLHF retains advantages.

\item {\bf Theory assumptions and proofs}
    \item[] Question: For each theoretical result, does the paper provide the full set of assumptions and a complete (and correct) proof?
    \item[] Answer: \answerNA{}
    \item[] Justification: The paper presents a conceptual taxonomy and empirical framework, not formal theoretical results requiring proofs.

\item {\bf Experimental result reproducibility}
    \item[] Question: Does the paper fully disclose all the information needed to reproduce the main experimental results of the paper to the extent that it affects the main claims and/or conclusions of the paper (regardless of whether the code and data are provided or not)?
    \item[] Answer: \answerYes{}
    \item[] Justification: Section~\ref{sec:experiments} specifies the three models (Qwen 2.5 3B Instruct, Llama 3.2 3B Instruct, Gemma 2 2B Instruct), both domains (RCPSP scheduling with OR-Tools CP-SAT verifier, MBPP with unit-test verifier), tier design, hardware (Apple M1, 32GB unified memory, MLX), training configuration matched across methods (6 iterations, identical LoRA rank, same temperature, same verifier as reward signal), and the iter-1-proxy baseline asymmetry for GRPO-LoRA. GRPO-LoRA hyperparameters (TRL/Unsloth defaults: LR $5{\times}10^{-6}$, $G{=}4$) are stated explicitly. Code is available at \url{https://anonymous.4open.science/r/isopro-836D}.

\item {\bf Open access to data and code}
    \item[] Question: Does the paper provide open access to the data and code, with sufficient instructions to faithfully reproduce the main experimental results, as described in supplemental material?
    \item[] Answer: \answerYes{}
    \item[] Justification: Code for \textsc{ISOPro}, the GRPO-LoRA baseline implementation, problem generators (scheduling and MBPP wrapper), verifiers, and evaluation scripts is available at \url{https://anonymous.4open.science/r/isopro-836D}. Problems are generated programmatically with fixed seeds.

\item {\bf Experimental setting/details}
    \item[] Question: Does the paper specify all the training and test details (e.g., data splits, hyperparameters, how they were chosen, type of optimizer) necessary to understand the results?
    \item[] Answer: \answerYes{}
    \item[] Justification: Section~\ref{sec:experiments} specifies tier design for both domains, training/eval split (T0--T4 train, T5 held out for scheduling; T0--T3 train, T4 held out for MBPP), LoRA configuration, temperature ($T{=}0.8$), rollout count, iterations (6), hardware, and runtime. Both methods use identical compute budgets and identical deterministic verifiers as the reward signal.

\item {\bf Experiment statistical significance}
    \item[] Question: Does the paper report error bars suitably and correctly defined or other appropriate information about the statistical significance of the experiments?
    \item[] Answer: \answerNo{}
    \item[] Justification: Cross-method results (Section~5.7) use a single seed (42) and 5 problems per tier; individual tier accuracies are directional estimates and the main empirical claims are stated at the cell-mean level. The original Qwen 2.5 3B / scheduling configuration was additionally run across 3 seeds for the ablation analysis (Section~5.8) with reported mean $\pm$ std. Extending 3-seed averaging to the cross-method matrix is a stated near-term replication priority. Acknowledged in Section~\ref{sec:limitations}.

\item {\bf Experiments compute resources}
    \item[] Question: For each experiment, does the paper provide sufficient information on the computer resources (type of compute workers, memory, time of execution) needed to reproduce the experiments?
    \item[] Answer: \answerYes{}
    \item[] Justification: All 12 (method, model, domain) cells run on an Apple M1 with 32GB unified memory using MLX, peak memory under 8GB across all configurations, no GPU required. Per-cell wall-clock times are reported in Section~5.6. Table~\ref{tab:arch} compares architectural memory requirements against RLHF and GRPO at frontier scale.

\item {\bf Code of ethics}
    \item[] Question: Does the research conducted in the paper conform, in every respect, with the NeurIPS Code of Ethics \url{https://neurips.cc/public/EthicsGuidelines}?
    \item[] Answer: \answerYes{}
    \item[] Justification: Uses publicly available models, no human subjects, scheduling and code-generation domains pose no dual-use risks. Safety implications discussed in Section~8.2.

\item {\bf Broader impacts}
    \item[] Question: Does the paper discuss both potential positive societal impacts and negative societal impacts of the work performed?
    \item[] Answer: \answerYes{}
    \item[] Justification: Section~8.2 discusses democratizing evaluation research and improving safety through better evaluation validity, including the safety dimension of accessibility.

\item {\bf Safeguards}
    \item[] Question: Does the paper describe safeguards that have been put in place for responsible release of data or models that have a high risk for misuse (e.g., pre-trained language models, image generators, or scraped datasets)?
    \item[] Answer: \answerNA{}
    \item[] Justification: The paper releases an evaluation framework, problem generators, verifiers, and a GRPO-LoRA baseline implementation, not a pre-trained model or scraped dataset. Task-specific LoRA adapters trained on scheduling and MBPP pose no misuse risk.

\item {\bf Licenses for existing assets}
    \item[] Question: Are the creators or original owners of assets (e.g., code, data, models), used in the paper, properly credited and are the license and terms of use explicitly mentioned and properly respected?
    \item[] Answer: \answerYes{}
    \item[] Justification: Qwen 2.5 3B Instruct (Apache 2.0), Llama 3.2 3B Instruct (Llama Community License), Gemma 2 2B Instruct (Gemma License), MLX, MBPP, and OR-Tools (Apache 2.0) are cited and used under their respective licenses.

\item {\bf New assets}
    \item[] Question: Are new assets introduced in the paper well documented and is the documentation provided alongside the assets?
    \item[] Answer: \answerYes{}
    \item[] Justification: \textsc{ISOPro} code, GRPO-LoRA baseline implementation, problem generators (scheduling and MBPP wrapper), and deterministic verifiers are released with documentation under an open-source license.

\item {\bf Crowdsourcing and research with human subjects}
    \item[] Question: For crowdsourcing experiments and research with human subjects, does the paper include the full text of instructions given to participants and screenshots, if applicable, as well as details about compensation (if any)?
    \item[] Answer: \answerNA{}
    \item[] Justification: No crowdsourcing or human subjects research.

\item {\bf Institutional review board (IRB) approvals or equivalent for research with human subjects}
    \item[] Question: Does the paper describe potential risks incurred by study participants, whether such risks were disclosed to the subjects, and whether Institutional Review Board (IRB) approvals (or an equivalent approval/review based on the requirements of your country or institution) were obtained?
    \item[] Answer: \answerNA{}
    \item[] Justification: No human subjects research.

\item {\bf Declaration of LLM usage}
    \item[] Question: Does the paper describe the usage of LLMs if it is an important, original, or non-standard component of the core methods in this research? Note that if the LLM is used only for writing, editing, or formatting purposes and does \emph{not} impact the core methodology, scientific rigor, or originality of the research, declaration is not required.
    \item[] Answer: \answerYes{}
    \item[] Justification: Qwen 2.5 3B Instruct, Llama 3.2 3B Instruct, and Gemma 2 2B Instruct are the base models for fine-tuning experiments, described in Section~\ref{sec:experiments}. The LLMs are the subject of evaluation.

\end{enumerate}

\end{document}